\documentclass[letterpaper]{article} 
\usepackage{aaai25}  
\usepackage{times}  
\usepackage{helvet}  
\usepackage{courier}  
\usepackage[hyphens]{url}  
\usepackage{graphicx} 
\usepackage{natbib}  
\usepackage{caption} 
\usepackage{algorithm}

\usepackage{algpseudocode}
\usepackage{tabularx}
\usepackage{booktabs}

\usepackage{amsmath}

\usepackage{newfloat}
\usepackage{listings}
\DeclareCaptionStyle{ruled}{labelfont=normalfont,labelsep=colon,strut=off} 
\floatstyle{ruled}
\newfloat{listing}{tb}{lst}{}
\floatname{listing}{Listing}
\title{Quantum-Classical Hybrid Molecular Autoencoder for Advancing Classical Decoding}
\author{
   Afrar Jahin\textsuperscript{\rm 1},
    Yi Pan\textsuperscript{\rm 2},\\
    Yingfeng Wang\textsuperscript{\rm 3*},
    Tianming Liu\textsuperscript{\rm 2*},
    Wei Zhang\textsuperscript{\rm 1*}
}
\affiliations {
    \textsuperscript{\rm 1}Department of Computer and Cyber Science, Augusta University, GA, USA\\
    \textsuperscript{\rm 2}School of Computing, The University of Georgia, Athens, GA, USA\\
     \textsuperscript{\rm 3}Department of Computer Science and Engineering, University of Tennessee at Chattanooga, Chattanooga, TN, USA\\
    
    ajahin@augusta.edu, ypan24@uga.edu, yingfeng-wang@utc.edu, tliu@uga.edu, wzhang2@augusta.edu\\
    {\rm *} Corresponding author
}

\usepackage{bibentry}

\begin{document}

\maketitle

\begin{abstract}

Although recent advances in quantum machine learning (QML) offer significant potential for enhancing generative models, particularly in molecular design, a large array of classical approaches still face challenges in achieving high fidelity and validity. In particular, the integration of QML with sequence-based tasks, such as Simplified Molecular Input Line Entry System (SMILES) string reconstruction, remains underexplored and usually suffers from fidelity degradation. In this work, we propose a hybrid quantum-classical architecture for SMILES reconstruction that integrates quantum encoding with classical sequence modeling to improve quantum fidelity and classical similarity. Our approach achieves a quantum fidelity of approximately \textbf{84\%} and a classical reconstruction similarity of \textbf{60\%}, surpassing existing quantum baselines. Our work lays a promising foundation for future QML applications, striking a balance between expressive quantum representations and classical sequence models and catalyzing broader research on quantum-aware sequence models for molecular and drug discovery.


\end{abstract}

%

\section{Introduction}

Molecular representation is a foundational task in computational chemistry, with broad applications in drug discovery~\cite{atz2021geometric, zeng2022accurate}.  Among the various representations, the SMILES strings provide a compact, interpretable way to encode molecular structures as sequences~\cite{zeng2022accurate}. However, accurately reconstructing and generating chemically valid SMILES strings remains a significant challenge due to their discrete syntax, structured dependencies, and context-sensitive rules. These challenges also hinder the application of quantum machine learning (QML), an emerging paradigm that leverages the quantum mechanical nature for advanced modeling~\cite{biamonte2017quantum, cerezo2022challenges, schuld2019quantum, pan2025molqae}, on molecular representations. 

In this work, we propose a quantum-classical hybrid molecular autoencoder (QCHMAE) to overcome the degradation of classical reconstruction (i.e., the reconstruction of SMILES strings). Notably, quantum autoencoders manipulate quantum states within a Hilbert space, allowing them to leverage quantum phenomena for more efficient information encoding~\cite{pan2025molqae}. The proposed framework integrates advancements of quantum circuits with the flexibility of classical decoders. In brief, our architecture employs a \textit{Word2Ket} embedding layer~\cite{panahi2019word2ket}, which encodes SMILES tokens into continuous quantum-inspired vectors using tensor-train decomposition~\cite{shi2023parallel, aksoy2024incremental}. These embedding results can be processed through a quantum autoencoder, and the results of quantum states are measured and fed into an attention-enhanced Long Short-Term Memory (LSTM) decoder~\cite{bahdanau2014neural} for SMILES reconstruction. In particular, our main contributions are:

\begin{itemize}
    \item We introduce a novel embeddings architecture, enabling quantum-inspired representation for SMILES strings ~\cite{shi2023parallel}. 
    
    \item We design an end-to-end pipeline that combines quantum encoding with an attention-enhanced LSTM decoder, effectively bridging quantum information processing with classical sequence generation.
    
    \item We provide empirical results that the proposed framework achieves a higher quantum fidelity (\textbf{84\%}) and classical similarity (\textbf{60\%}). 
    
    \item We propose a novel loss function that jointly optimizes quantum fidelity, trash deviation loss, and sequence-level similarity. 

\end{itemize}

\begin{figure*}[ht]
    \centering
    \includegraphics[width=1\textwidth]{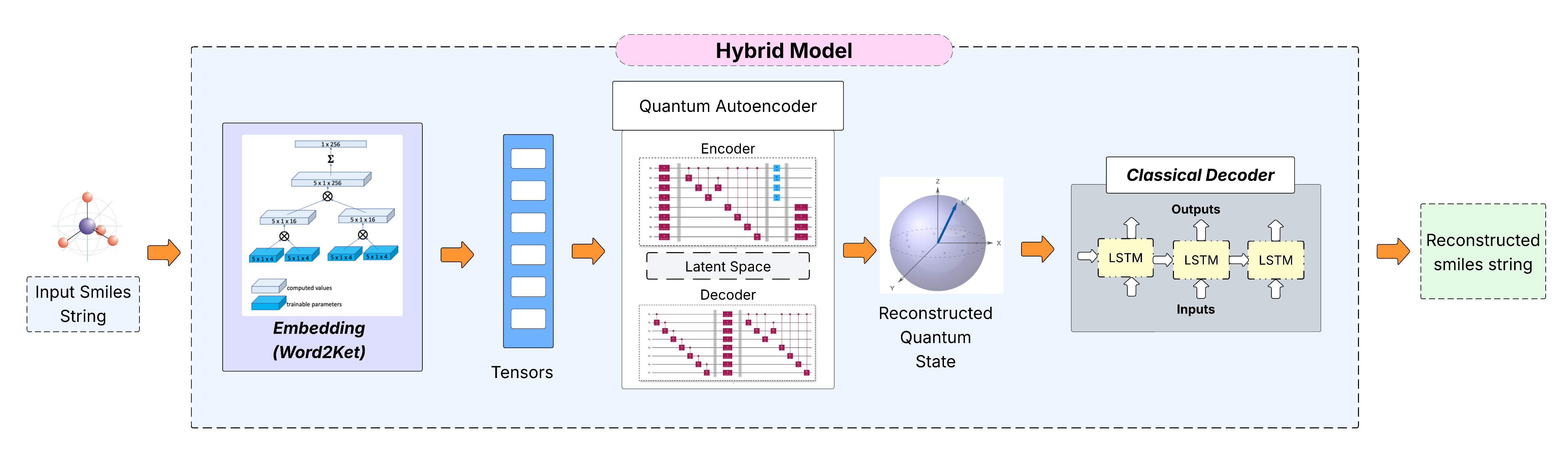}
    \caption{Overview of the proposed hybrid quantum-classical architecture for SMILES reconstruction.}
    \label{fig:model_architecture}
\end{figure*}

\noindent \textbf{Related work:} Traditional machine learning models for molecular exploration have largely relied on engineered descriptors, graph neural networks (GNNs), or SMILES-based sequence models to capture the chemical and structural properties of molecules~\cite{gilmer2017neural, li2022deep, wu2018moleculenet}. However, these models struggle to encode quantum mechanical characteristics, generalize to out-of-distribution molecules, or accommodate conformational flexibility~\cite{yang2019analyzing, schwaller2019molecular, wang2019smiles}. 

In addition, recent advances in QML have enabled the development of quantum circuits and hybrid models for molecular data analytics~\cite{schuld2019quantum, cerezo2021variational, albrecht2023quantum, pan2025molqae}. Quantum autoencoders have been proposed to explore quantum states, showing promise for molecular applications~\cite{bondarenko2020quantum, pan2025molqae}. For example, MolQAE~\cite{pan2025molqae} is a recently proposed quantum-classical framework designed for the efficient compression of molecular representations. 
Furthermore, hybrid quantum-classical frameworks integrate quantum representation with classical neural models to leverage the strengths of both paradigms~\cite{biamonte2017quantum, lloyd2020quantum}. Such architectures have been explored for tasks like molecular property prediction, quantum chemistry simulations, and molecular generation~\cite{cao2019quantum, farhi2014quantum, huang2021power, gircha2023hybrid}. 

Notably, our proposed approach extends prior work by incorporating a quantum autoencoder for latent representation learning, followed by an embedding layer and an attention-enhanced LSTM decoder to enhance the reconstruction of SMILES strings.

\section{Methodology}

Our proposed architecture (Figure~\ref{fig:model_architecture}) integrates quantum-inspired embeddings, quantum autoencoding, and classical sequence decoding to reconstruct SMILES strings. The framework comprises three core components: (1) an embedding layer, (2) a quantum autoencoder, and (3) an attention-enhanced LSTM decoder. As depicted in Figure~\ref{fig:model_architecture}, the input SMILES string is first mapped into a quantum-inspired Hilbert space using a \textit{Word2Ket} embedding. The resulting quantum state is then processed by the quantum autoencoder to learn a compressed latent representation. Finally, the decoded quantum state is passed through an attention-enhanced LSTM to generate the reconstructed molecular sequence. The framework is optimized using a hybrid loss function that jointly balances quantum fidelity and classical sequence similarity, enabling effective integration of quantum and classical learning paradigms.

\subsection{Embedding Layer}

Initially, to encode input SMILES strings into a suitable format for quantum processing, we employ the \textit{Word2Ket} embedding framework~\cite{panahi2019word2ket}. Each tokenized SMILES sequence $\mathbf{x} = [x_1, x_2, \dots, x_L]$ is mapped into a high-dimensional Hilbert space through tensor-train decomposition~\cite{shi2023parallel,aksoy2024incremental}. Specifically, the embedding of a sequence is formulated as follows:
\begin{equation}
    \mathbf{z} = \mathrm{Word2Ket}(\mathbf{x}) = \prod_{i=1}^{L} \mathbf{E}_{x_i}
\end{equation}
where $\mathbf{E}_{x_i}$ denotes the trainable tensor assigned to token $x_i$, and the product is a tensor contraction over shared indices. This approach enables the model to capture quantum-like entanglement patterns between distant tokens with efficient memory scaling, outperforming traditional embedding methods in expressivity~\cite{panahi2019word2ket}.

\subsection{Quantum Autoencoder}

The embedded representation $\mathbf{z}$ is then processed by a parameterized quantum circuit (PQC) serving as a quantum autoencoder~\cite{pan2025molqae, bondarenko2020quantum}. The PQC encodes the input vector into a quantum state $|\psi_\mathbf{z}\rangle$ using a sequence of single- and multi-qubit gates:
\begin{equation}
    |\psi_\mathbf{z}\rangle = \mathcal{U}_\theta(\mathbf{z}) |0\rangle^{\otimes n}
\end{equation}
where $\mathcal{U}_\theta$ denotes the unitary evolution parameterized by $\theta$, conditioned on the input embedding. The quantum autoencoder compresses essential information into a lower-dimensional latent space, while trash qubits are encouraged to remain in the ground state. The fidelity between the initial and reconstructed quantum states is computed as:
\begin{equation}
    \mathcal{L}_{\mathrm{fidelity}} = 1 - |\langle \psi_\mathbf{z}^{\mathrm{in}} | \psi_\mathbf{z}^{\mathrm{out}} \rangle|^2
\end{equation}
ensuring quantum information is preserved during encoding and decoding~\cite{romero2017quantum, bondarenko2020quantum}.

\begin{table}[ht!]
\centering
\caption{Experimental Setup for Hybrid Quantum-Classical SMILES Autoencoder}
\begin{tabular}{ll}
\toprule
\textbf{Component}         & \textbf{Value / Description} \\

\midrule
\textbf{Quantum Encoder}   &  \\
Number of encoder qubits   & 8 \\
Number of latent qubits    & 5 \\
QAE layers                 & 5 \\
Trash qubits               & 4 \\
Entanglement topology      & CRZ gates \\
\midrule
\textbf{Classical Decoder} &   \\
Hidden dimension           & 252 \\
Decoder layers             & 4 \\
Attention heads            & 8 \\
\midrule
\textbf{Training}          & \\
Batch size                 & 1024  \\
Learning rate              & $1 \times 10^{-6}$ \\
Optimizer                  & Adam \\
Epochs                     & 50 \\

\bottomrule
\end{tabular}
\label{tab:experimental-setup}
\end{table}

\subsection{Attention-Enhanced Classical Decoder}

The quantum state is measured and projected back into a classical latent vector $\hat{\mathbf{z}}$, which is then provided as input to an attention-enhanced LSTM decoder. This decoder models the sequential dependencies in the SMILES string, leveraging both the quantum-processed features and a self-attention mechanism:
\begin{align}
    \mathbf{h}_t, \mathbf{c}_t &= \mathrm{LSTM}([\hat{\mathbf{z}}, \mathbf{y}_{<t}], \mathbf{h}_{t-1}, \mathbf{c}_{t-1}) \\
    \mathbf{a}_t &= \mathrm{Attention}(\mathbf{h}_t, \mathbf{H}_{\mathrm{enc}})
\end{align}
where $\mathbf{y}_{<t}$ denotes the previously generated tokens and $\mathbf{H}_{\mathrm{enc}}$ represents the encoded input states. The final output distribution over SMILES tokens is obtained via a linear projection:
\begin{equation}
    P(y_t \mid y_{<t}, \hat{\mathbf{z}}) = \mathrm{Softmax}(\mathbf{W}_o [\mathbf{h}_t ; \mathbf{a}_t] + \mathbf{b}_o)
\end{equation}
where $[\cdot ; \cdot]$ denotes concatenation.

To quantify the accuracy of classical similarity (i.e., SMILES string reconstruction), we employ Levenshtein similarity as one of our primary evaluation metrics~\cite{janardhana2024efficient}. This similarity yields a value within the range $[0, 1]$, with higher scores indicating a strong similarity between the original input and reconstructed sequences. 
The classical similarity is illustrated as follows:
\begin{equation}
    \text{Levenshtein Similarity} = 1 - \frac{d(A, B)}{\max(|A|, |B|)}
\end{equation}
where $d(A, B)$ denotes the Levenshtein distance~\cite{janardhana2024efficient} between the original sequence $A$ and the reconstructed sequence $B$, and $|A|$ and $|B|$ are their respective lengths. 

\subsection{Quantum-Classical Hybrid Molecular Autoencoder}

The proposed hybrid molecular autoencoder is optimized end-to-end in an unsupervised manner~\cite{bengio2012unsupervised} using mini-batch stochastic gradient descent, jointly updating both the quantum encoder and classical sequence decoder. In each training iteration, input SMILES strings are tokenized and embedded, then projected into a quantum feature space via the \textit{Word2Ket} embedding followed by a classical transformation. The quantum autoencoder encodes these features into latent quantum states, which are subsequently measured and converted into classical vectors for downstream decoding. An attention-enhanced LSTM decoder reconstructs the SMILES sequence, effectively bridging quantum representations with classical sequence generation. The learning objective is defined as a weighted sum of multiple loss components: quantum state fidelity, cross-entropy for sequence reconstruction, Levenshtein similarity for SMILES string alignment~\cite{janardhana2024efficient}, and a penalty for trash qubit deviation~\cite{philips2022universal}. This composite loss ensures the preservation of quantum information. To further stabilize this process, teacher forcing with scheduled sampling~\cite{bolboacua2023performance} is employed, guiding the decoder with ground truth sequences during early stages.

The detailed procedure is summarized in Algorithm~\ref{alg:hybrid-training}.
\begin{algorithm}[h]
\caption{Quantum-Classical Hybrid Framework}
\label{alg:hybrid-training}
\begin{algorithmic}[1]
\State \textbf{Input:} Tokenized SMILES dataset $\mathcal{D}$; embedding function $\mathrm{Embed}$; quantum encoder $\mathcal{U}_\theta$; LSTM decoder $\mathrm{Decoder}$; loss weights $\lambda_1, \dots, \lambda_4$; teacher forcing schedule $\alpha(t)$
\For{each epoch $t = 1$ to $T$}
    \For{each mini-batch $(x, y)$ in $\mathcal{D}$}
        \State $z \gets \mathrm{Embed}(x)$ \Comment{Word2Ket embedding}
        \State $|\psi_z\rangle \gets \mathcal{U}_\theta(z)$ \Comment{Quantum encoding}
        \State $\hat{z} \gets \mathrm{Measure}(|\psi_z\rangle)$ \Comment{Measurement (classical projection)}
        \State $\hat{y} \gets \mathrm{Decoder}(\hat{z}, y, \alpha(t))$ \Comment{LSTM decoding with scheduled sampling}
        \State Compute $\mathcal{L}_{\text{fidelity}}$, $\mathcal{L}_{\text{CE}}$, $\mathcal{L}_{\text{SMILES}}$, $\mathcal{L}_{\text{trash}}$
        \State $\mathcal{L}_{\text{total}} \gets \lambda_1 \mathcal{L}_{\text{fidelity}} + \lambda_2 \mathcal{L}_{\text{CE}} + \lambda_3 \mathcal{L}_{\text{SMILES}} + \lambda_4 \mathcal{L}_{\text{trash}}$
        \State Update model parameters via backpropagation
    \EndFor
\EndFor
\end{algorithmic}
\end{algorithm}

\section{Experiments and Discussion}

\subsection{Dataset and Experimental Setup}

In experiments, we employed the QM9 dataset~\cite{blum2009970}, which contains roughly 134,000 organic molecules, each with no more than nine heavy atoms. Prior to training, the dataset was processed by canonicalizing the SMILES representations with RDKit~\cite{rdkit}, eliminating duplicates, and excluding invalid molecular entries. The experimental details are summarized in Table~\ref{tab:experimental-setup}.

\subsection{Training and Results}

\begin{figure}[H]
\centering
\includegraphics[width=1.0\columnwidth]{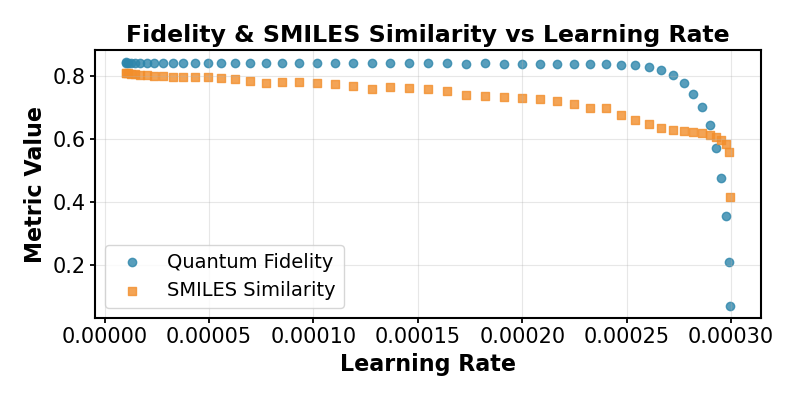} 

\caption{Quantum Fidelity and SMILES Similarity plotted against the learning rate across training epochs.}
\label{fig2}
\end{figure}

\begin{figure}[t]
\centering
\includegraphics[width=1\columnwidth]{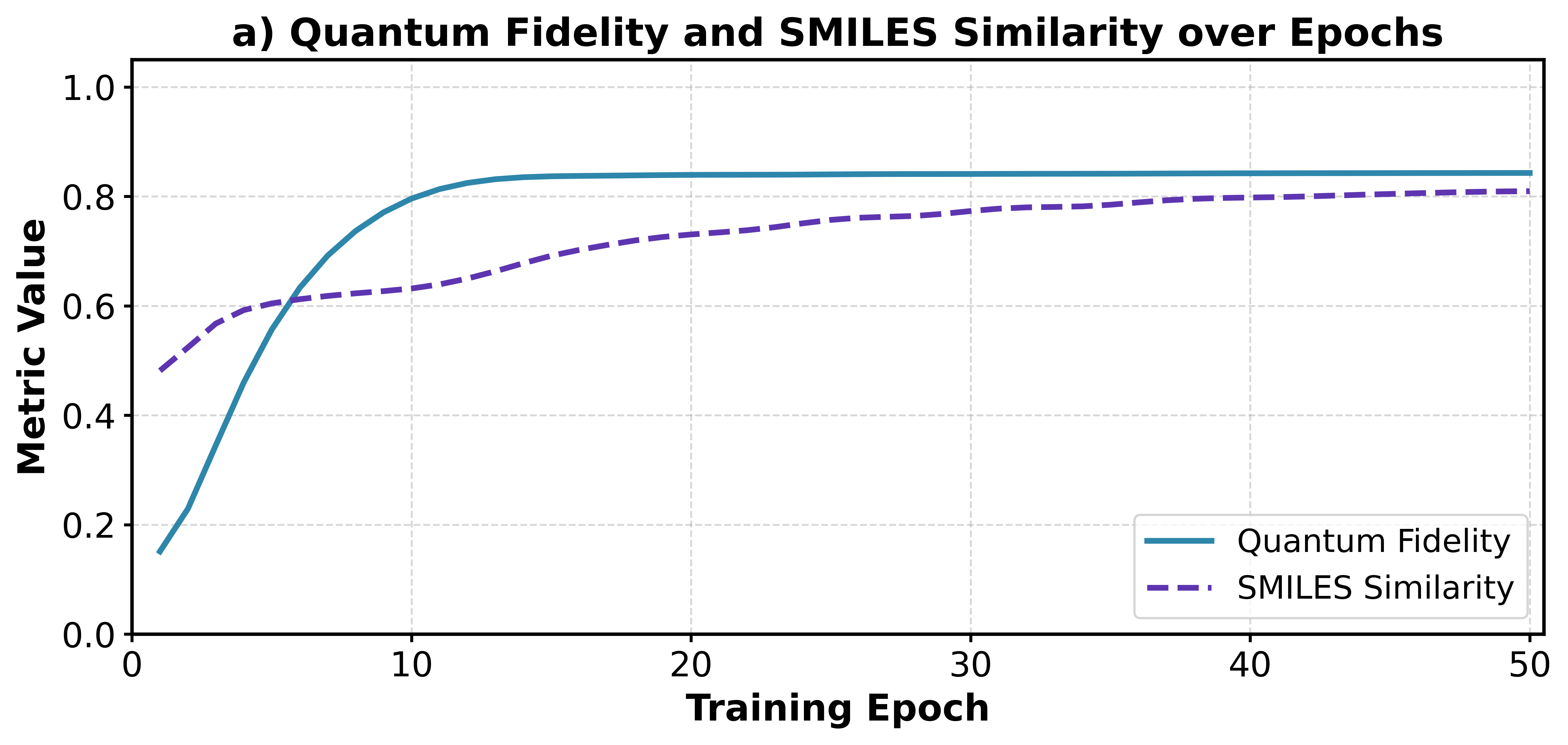} 
\includegraphics[width=1\columnwidth]{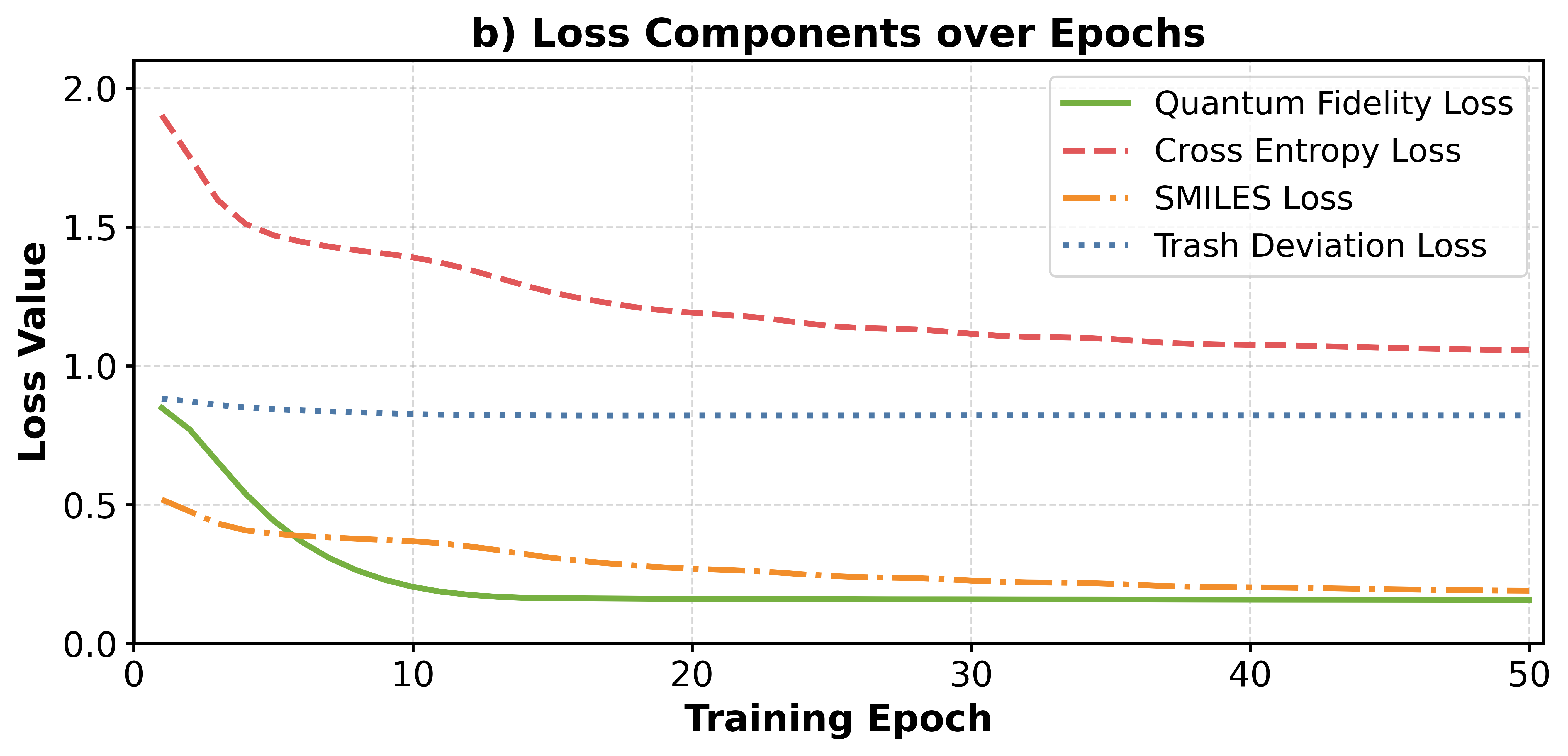} 

\caption{(a) Quantum Fidelity and SMILES Similarity plotted over training epochs. 
(b) Breakdown of loss components during training. Quantum fidelity loss, cross-entropy loss, SMILES loss, and trash deviation loss are shown with smoothed curves.} 
\label{fig3}
\end{figure}

The results of training the proposed hybrid model are illustrated in Figures~\ref{fig2} and~\ref{fig3}. The training focuses on unlabeled data without testing, such as k-fold validation~\cite{bengio2012unsupervised}. Overall, the results demonstrate that quantum fidelity increases steeply and stabilizes above 80\%, while SMILES similarity (i.e., classical similarity) improves around 60\%. 
In particular, all loss curves decrease sharply during the initial training phase and gradually converge to stable values.  
These trends demonstrate that enhancements in quantum representation directly contribute to improved downstream molecular sequence reconstruction within the hybrid architecture.

Furthermore, Figure~\ref{fig2} illustrates the relationship between the learning rate and the performance throughout training. As the learning rate decays under the CosineAnnealingLR scheduler, both quantum fidelity and SMILES similarity remain consistently high. This trend highlights the effectiveness of the scheduling strategy, facilitating rapid optimization in the early stages of training and enabling fine-grained tuning in later epochs. 
Moreover, Figure~\ref{fig3} presents the quantum fidelity, SMILES similarity (i.e., classical similarity), and 4 loss components within 50 epochs. These results highlight the model's capability to quickly learn quantum representations and progressively enhance classical reconstruction accuracy.
\begin{table}[H]
    \centering
    \caption{QML Models performance comparison under classical decoding}
    \label{table:performance_comparison}
    \begin{tabularx}{\linewidth}{>{\centering\arraybackslash}X >{\centering\arraybackslash}X >{\centering\arraybackslash}X >{\centering\arraybackslash}X >{\centering\arraybackslash}X }
        \toprule
        \textbf{QML Models} & \textbf{Embeddi-ng type}  & \textbf{Quantum Fidelity(\%)} & \textbf{Trash Deviation(\%)} & \textbf{Classical Similarity (\%)} \\ 
        \midrule
        MolQAE  & ASCII & 77 & 90 & N/A \\
        \addlinespace
       QCHMAE & Word2ket & \textbf{84} & \textbf{82} & \textbf{60} \\
        \bottomrule
    \end{tabularx}
\end{table}

Besides, we compare the SMILES reconstruction using the only existing quantum model, MolQAE. Table~\ref{table:performance_comparison} provides a quantitative evaluation between our QCHMAE and the prior MolQAE~\cite{pan2025molqae} without a classical decoder. In contrast, our hybrid model with a \textit{Word2Ket}~\cite{panahi2019word2ket} embedding substantially outperforms, achieving higher quantum fidelity, a lower trash deviation, and a marked improvement in classical similarity of SMILES strings reconstruction as well. 

\subsection{Discussion}

In terms of model performance, the hybrid architecture achieves a quantum fidelity of \textbf{84\%}, indicating that the quantum autoencoder effectively preserves latent state information through the encoding and decoding processes. The average classical similarity~\cite{janardhana2024efficient} between original and reconstructed SMILES strings is \textbf{60\%}, reflecting an accurate reconstruction at the token-level sequence. Additionally, a trash deviation score of 82\% confirms that the non-latent (“trash”) qubits~\cite{philips2022universal} are successfully driven toward the zero state, validating the effectiveness of the quantum compression mechanism. In summary, these results underscore the strengths of the proposed hybrid approach. 

\section{Conclusion}

In this work, we integrated a quantum autoencoder with classical sequence modeling to significantly improve quantum fidelity and classical similarity. The results showcase that our framework can improve quantum fidelity to \textbf{84\%} and achieve the classical similarity as \textbf{60\%}. This surpasses prior MolQAE, highlighting the value of combining quantum feature processing with classical decoding. 
Notably, our proposed framework also faces challenges. The improvement of quantum fidelity does not always lead to an improvement in classical similarity. This highlights a fundamental challenge in hybrid quantum-classical architectures. Overall, this work illustrates a viable path forward for integrating quantum learning modules into molecular generation pipelines. With improvements in quantum training, such hybrid models could become increasingly practical and impactful for quantum-enhanced molecular representation and drug discovery.

\bibliography{aaai25}

\end{document}